\documentclass{article}

\usepackage[preprint]{neurips_2021}

\usepackage[utf8]{inputenc} 
\usepackage[T1]{fontenc}    
\usepackage{hyperref}       
\usepackage{url}            
\usepackage{booktabs}       
\usepackage{amsfonts}       
\usepackage{mathtools}
\usepackage{nicefrac}       
\usepackage{microtype}      
\usepackage{xcolor}         
\usepackage{natbib}
\setcitestyle{numbers,square}

\usepackage{graphicx}
\usepackage{subcaption} 
\usepackage{color}
\usepackage{paralist}
\usepackage{comment}
\usepackage{multirow}
\usepackage{url}
\usepackage{amsmath,amssymb,framed}
\usepackage[linesnumbered,ruled]{algorithm2e}
\usepackage{balance}
\usepackage{epstopdf}
\usepackage{epsfig}
\usepackage{hyperref}
\usepackage[flushleft]{threeparttable}
\usepackage{paralist}
\usepackage{caption}
\usepackage{setspace}
\usepackage{authblk}
\usepackage{threeparttable}

\makeatletter
\def\BState{\State\hskip-\ALG@thistlm}
\makeatother
\usepackage{bbm}
\usepackage{booktabs}
\usepackage{array}
\newcolumntype{L}{>{$}l<{$}}
\newcolumntype{C}{>{$}c<{$}}
\newcolumntype{R}{>{$}r<{$}}

\DeclareMathOperator*{\argmin}{arg\,min}

\title{Time Series Anomaly Detection with label-free Model Selection}

\author{Deokwoo Jung}
\author{Nandini Ramanan}
\author{Mehrnaz Amjadi}
\author{\\Sankeerth Rao Karingula}
\author{Jake Taylor}
\author{Claudionor Nunes Coelho Jr}

\affil{ADVANCED APPLIED AI RESEARCH \\
PALO ALTO NETWORKS, USA 
\authorcr
  {\{\tt djung,nramanan,mamjadi,skaringula,jataylor,ccoelho\}@paloaltonetworks.com}
\date{May 2021}

}

\begin{document}
\newlength{\commentWidth}
\setlength{\commentWidth}{7cm}
\newcommand{\atcp}[1]{\tcp*[r]{\makebox[\commentWidth]{#1\hfill}}}

\maketitle

\begin{abstract}
Anomaly detection for time-series data becomes an essential task for many data-driven applications fueled with an abundance of data and out-of-the-box machine-learning algorithms. In many real-world settings, developing a reliable anomaly model is highly challenging due to insufficient anomaly labels and the prohibitively expensive cost of obtaining anomaly examples. 
It imposes a significant bottleneck to evaluate model quality for model selection and parameter tuning reliably.
As a result,  many existing anomaly detection algorithms fail to show their promised performance after deployment. 

In this paper, we propose \texttt{LaF-AD}, a novel anomaly detection algorithm with label-free model selection for unlabeled times-series data.
Our proposed algorithm performs a fully unsupervised ensemble learning across a large number of candidate parametric models. We develop a model variance metric that quantifies the sensitivity of anomaly probability with a bootstrapping method. 
Then it makes a collective decision for anomaly events by model learners using the model variance. 
Our algorithm is easily parallelizable, more robust for ill-conditioned and seasonal data, and highly scalable for a large number of anomaly models.  We evaluate our algorithm against other state-of-the-art methods on a synthetic domain and a benchmark public data set. 

\end{abstract}
\section{Introduction} \label{sec:intro}

As Internet of Things (IoT)-enabled devices and cloud infrastructure are increasingly available for critical data-driven applications such as cloud-based security services, there is an explosion of time series data that is now available \cite{singh2017cloud}. There is an immense opportunity emerging to utilize time series data to extract actionable information for operations and maximize operational reliability and security. In particular, detecting anomalies during operations is critical for its immediate application to reduce unplanned service downtime and improve site reliability in cloud-based applications. 
 A single hour of unplanned downtime or security breach in high-value/high-volume data applications can lead to millions of dollars operational losses \cite{kirschen2003computing}.  Despite the need, it is highly challenging to develop anomaly detection or compare different models in an industrial setting that is reliably scalable for various applications due to the lack of data labels, system information,  and poor data quality.

In many real-world data-driven applications, anomaly detection often does not perform as expected due to the following reasons.  Firstly, labeled data sets are rarely available, expensive, and often impractical to obtain in significant quantities. Furthermore, there is usually no a priori knowledge about which points are normal or abnormal. There also can be ambiguity in the definition of an anomaly. As a result, it needs to be learned from data without normal or abnormal examples (i.e., unsupervised learning). Inevitably, it imposes a significant limitation on optimizing model parameters since no validation dataset is available to test out-of-sample errors.  In particular, using a more complex model (i.e., more parameters) often causes significant out-of-sample errors rendering the model highly unreliable over unseen data sets. 

Secondly, time-series data from IoT devices and cloud infrastructures is often ill-conditioned, meaning that it contains a large number of missing, corrupted,  and noisy training samples. Such data characteristic becomes more eminent for a higher dimensional sensor data as the average noise level, and missing value event per a training sample grows with its dimensionality. Learning reliable anomaly models from noisy data without labels is a highly challenging problem as there is no way to distinguish noise from a signal. Using a complex model tends to inadvertently employ its larger number of degrees of freedom in model parameters to fit the noise, which causes a significant model overfitting resulting in large out-of-sample errors.

Finally, learned anomaly models must be easily interpretable and trusted by users for advanced industrial analytics applications.  Without the ability to explain why anomalies are detected, anomaly models are unlikely to be adopted for real industrial applications regardless of their accuracy. Anomalies that arise in industrial applications ultimately need to be explainable to field engineers or operators for further corrective or preventive actions. Hence, it is essential for a learning algorithm to obtain interpretable features in anomaly models from high-dimensional complex time series data.

In this paper, we propose a novel anomaly detection algorithm, \underline{La}bel-\underline{f}ree \underline{A}nomaly \underline{D}etection or \texttt{LaF-AD}, for unlabeled times-series data. Proposed \texttt{LaF-AD} performs a fully unsupervised ensemble learning across a large number of candidate parametric models. 
We develop a model variance metric that quantifies the sensitivity of anomaly probability with a bootstrapping method. Then it makes a collective decision for anomaly events by model learners using the model variance. 
Our algorithm can be easily parallelizable, more robust for ill-conditioned data, and highly scalable for a large number of anomaly models.  More importantly, it is highly interpretable to many feature sensors compared to other conventional approaches that use multivariate complex anomaly models. We evaluate our algorithm against synthetic and real-world public datasets.

The rest of the paper is organized as follows. Section \ref{sec:related} reviews the previous studies on data analytical techniques.
Section \ref{sec:anomlay_detection} describes our proposed novel anomaly detection algorithm. 
Section \ref{sec:expr} presents experimental results towards validation of the algorithm. 
Section \ref{sec:conclude} offers a summary and some concluding remarks.

\section{Related Work} \label{sec:related}
Anomaly detection has been studied and applied to time series data for decades~\cite{braei2020anomaly}. Anomaly detection is cast into the three paradigms based on the nature of the data capturing anomalies; supervised, semi-supervised and unsupervised approaches~\cite{Aggarwal16}. In several scenarios, we can leverage previously known anomalies in the domain to label the data as non-anomalous or anomalous, paving the way for supervised or semi-supervised outlier detection. However, supervised approaches require high volumes of labeled instances of both non-anomalous and anomalous data to learn robust hypotheses from, making it impractical for real-world problems~\cite{GornitzKRB14}. Semi-supervised algorithms are often derived from supervised anomaly detection methods by employing a bias term to handle the unlabeled data.~\cite{vapnik1999overview,sindhwani2005beyond}.

More recently, unsupervised methods are preferred, which is also the concentration of this work. Most of the research under unsupervised anomaly detection can be categorized into statistical methods, distance-based, ensemble-based, and reconstruction-based techniques. Our proposed methodology combines concepts from each of these categories.

\noindent \textbf{Statistical methods:} Seminal probabilistic approaches focused on the marginal likelihood estimates $p_\theta(\mathbf{X})$ of the data generation model to score the outliers~\cite{Goernitz2015,Yongjie2015,Liang2011,Junlei2009}. A probabilistic algorithm by Dempster et al. employed the Gaussian Mixture Model (GMM) to fit a number of Gaussians to the data and employed expectation–maximization (EM) algorithm to estimate the parameters~\cite{Dempster1977}. Blei \& Jordan built upon this by using Dirichlet Process nonparametric models for adaptively determining the number of clusters based on the complexity of the data. Earlier work by Scholkopf et al. proposes One-class SVMs, nonparametric models in which an underlying probability density is assumed (such as fitting GMMs)~\cite{scholkopf2001}. One of the drawbacks of the statistical methods is that they can only detect anomalies w.r.t the global data distribution~\cite{song2007}. Generative models also suffer as they are approximated as a simple linear model with conjugate priors to derive an analytical solution, due to the complexity in computing marginal likelihoods $p_\theta(\mathbf{X}) = \int p_\theta(\mathbf{z})p_\theta(\mathbf{X}|\mathbf{z})d\mathbf{z}$, that often requires computing very high dimensional integrals. 

\noindent \textbf{Distance based methods:} The most popular method under the umbrella of distance-based anomaly detection is proposed by Knorr et al. wherein a point p is considered an $(\pi, \epsilon)$-outlier if at most $\pi\%$ of all points are less than $\epsilon$ away from p~\cite{knorr2000}. More formally, $|\{q\in\mathcal{D} \,\,\,\,\, dist(p,q)<\epsilon\}| \geq \pi.n$. This approach was later extended by Ramaswamy et al., who proposed a highly efficient way to calculate the anomaly score by computing the distance to the $k^{th}$ Nearest Neighbour of a point~\cite{ramaswamy2000efficient}. Then, thresholds are used to classify a data point as anomalous or not. Another faster and scalable variant of this effort is proposed by Angiulli $\&$ Pizzuti where a  KNN based algorithm that takes the aggregate distance of point $p$ overall its $k$ nearest neighbors as an anomaly score~\cite{Fabrizio2002}, i.e., 
\begin{align}
    NN(p;k,\mathcal{D}) =  \frac{1}{|\eta_k(p)|}\sum_{q\in \eta_k(p)} dist(p,q)
\end{align}
where $\eta_k(p)$ returns the $k$ nearest neighbours of $p$ in $\{\mathbf{D}\setminus\{p\}\}$. Another popular method, local outlier factor (LOF) by Breunig et al. computes the outlier score as a ratio of the average of densities of the k nearest neighbors to the density of the instance itself~\cite{breunig2000lof}.  
Both KNN and LOF have been shown to do exceedingly well compared with state-of-the-art publicly available anomaly detection methods in real problems. Still, both techniques do not scale well with large high-dimensional and seasonal data. ~\cite{campos2016evaluation}.

\noindent \textbf{Ensemble based methods:} The first ensemble learning approach to outlier detection runs on LOF when they are learned with different sets of hyperparameters such that the resultant combination is the anomaly scores~\cite{Xu_2019}. Isolation Forest (IF) is another ensemble-based algorithm that builds a forest of random binary trees such that anomalous instances have short average
path lengths on the trees~\cite{liu2008isolation,liu2012isolation,Hariri_2021}. 

\noindent \textbf{Reconstruction based methods:} Recent advent in anomaly detection compute synthetic reconstruction of the data. These approaches work because once projected to a lower-dimensional space, anomalies lose information, which prevents them from being reconstructed effectively. This leads to a higher reconstruction error for anomalous points. We compute the difference between an observed value and its reconstruction as $\frac{1}{\eta}\sum_{i=1}^\eta ||p_i-p'_i||_2^2 $, where $p_i$ and $p'_i|$ are the i-th features and its reconstruction, respectively. The well-known principal component analysis (PCA) based anomaly detection by Rousseeuw et al. can be employed to reconstruct the data but only allows for linear reconstruction~\cite{rousseeuw2018anomaly}. Kernel PCA is a non-linear enhancement of PCA by Hoffmann et al. for outlier detection where data is mapped into the features space using the kernel trick~\cite{hoffmann2007kernel}. We shift our focus to another reconstruction-based method which we use in our work, autoencoders (AE), one of the most widely used deep learning methods for anomaly detection~\cite{wen2017new,chen2017outlier}. Alternatively, researchers have proposed an RNN- and LSTM-based autoencoders model for anomaly detection in time series~\cite{cho2014learning,principi2017acoustic}. Malhotra et al. were the first to apply the LSTM-based encoder-and-decoder for anomaly detection to time series data which demonstrated better generalization capability than other prediction-based method~\cite{malhotra2016lstm}. 

In \texttt{LaF-AD} we aim to provide an explainable approach by combining reconstruction-based method LSTM-AE and ensemble methods for unsupervised anomaly detection, incorporating the advantages of both models. In addition, we propose a model variance metric that quantifies the sensitivity of anomaly probability with a bootstrapping method. 

\section{Learning Label-free Anomaly Model} \label{sec:anomlay_detection}
This section formally describes algorithms for learning the anomaly models and provides a detailed explanation of our proposed anomaly detection method.
We use the following simplified matrix notations.
We use the notation $\mathbf{A}_{n \times m} = [a_{ij}]^{i_n, j_m}_{i,j =i_1, j_1} $ for  $n \times m $ matrix of $\mathbf{A}$ where  $i_1  \leq i \leq i_n$ and $j_1  \leq j \leq j_m$. 
The $i$th row vector and the $j$th column vector of $\mathbf{A}$ are denoted by $\mathbf{A}_{i.}$  and $\mathbf{A}_{.j} $
The inner product of  $\mathbf{a}$ and $\mathbf{b}$ is denoted by $\langle \mathbf{a},\mathbf{b} \rangle$. The expectation of random variable $a$ is denoted by $\bar{a}$.

\subsection{Algorithm Overview}
\begin{figure*}
	\centering
	\includegraphics[width=1.00\textwidth,trim={0cm 6cm 0cm 0cm}]{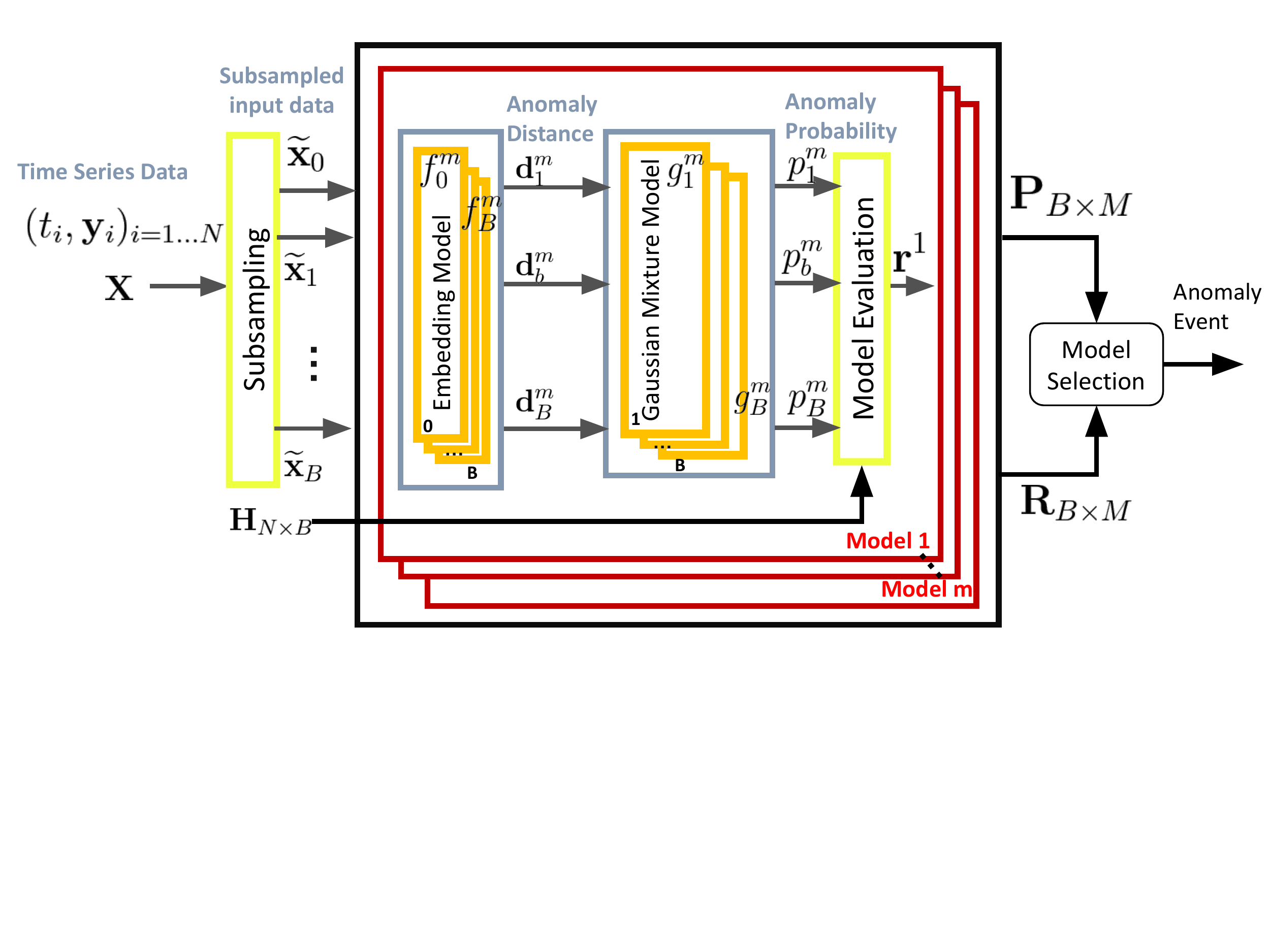}
	\caption{Algorithm Overview}
	\label{fig:system_a}
		\vspace{-12pt}
\end{figure*}
Let us formally describe our problem formulation and overall approach of our algorithm. Suppose that we have $N$ samples of time series data set for training 
$\mathbf{X}_{train} = ((t_n, \mathbf{y}_n))^N_{n=1}$ where $t_n$ and $\mathbf{y}_n$ are a time stamp and time series data at sample index $n$. 
Similarly, let denote $\mathbf{X}_{val} = ((t_{n+1}, \mathbf{y}_{n+k}))^K_{k=N+1}$ for validation data set for model selection. Let define $\mathbf{x}_i =(t_i, \mathbf{y}_i) $ and $\mathbf{x}$ denote a random variable of $\mathbf{x}_i$.

Then anomaly detection algorithm $\mathcal{A}$ is defined by 
$\mathcal{A}: \mathbf{x}_i \mapsto  v_i \in \{0,1\}$ where $v_i=1$ represents an anomaly event and $0$ for a normal event.
The anomaly event is determined by setting a threshold $0.5$ to anomaly probability $p_i$ such that $v_i = \mathbbm{1}_{[p_i >0.5]}$ where $\mathbbm{1}$ is an indicator function.
We aim to estimate $\mathbbm{E}[\sigma^2_v|\mathcal{A}]$, an expected out-of-sample variance for $v_i$ from $\mathbf{X}_{val}$ given an anomaly algorithm $\mathcal{A}$ where $\sigma^2_v = Var(v_i)$.

For ensemble method, we assume that $M$ anomaly models $f_1\cdots f_M$ are learnt from their respective algorithm $\mathcal{A}_m$ and a training data set $\mathbf{X}_{train}$.
Hence, anomaly ensemble model $\{f_m\}^M_{m =1}$ is learnt by 
$\mathcal{A}_m: \mathbf{X}_{train}  \mapsto f^m $ and $f^m : \mathbf{x}_i \mapsto v_i$.
The the expected variance of anomaly model $m$ is formally described as following,
\begin{equation}\label{eqn:model_var}
\mu^m_{\sigma} \coloneqq \mathbbm{E}_{\mathbf{x}} [Var (f^m(\mathbf{x}))].
\end{equation}

Let use a simplified notation  $v^m \coloneqq f^m(\mathbf{x})$. 
It is easily shown that  $0 \leq \mu^m_{\sigma} \leq 0.25$ since $Var(v^m) = \bar{v}^m (1-\bar{v}^m)$ and  $0 \leq \bar{v}^m \leq 1$.
Note that the model variance $\mu^m_{\sigma} \to 0.25$ as out-of-sample anomaly prediction (i.e., predictions on $\mathbf{X}_{val}$) becomes highly unstable $\bar{v}^m \to 0.5$.
Conversely, for more reliable anomaly prediction we have $\bar{v}^m \to 1$ or $0$ (i.e., $\mu^m_{\sigma} =0$). 

To aggregate the anomaly models we derive ensemble weights of models 
$ w^m_{esm} = \frac{1-4\mu^m_{\sigma}} {\sum^M_{m=1} (1-4\mu^m_{\sigma})} $ that sets a zero weight for $\mu^m_{\sigma} = 0.25$.
The ensemble model output $v_i^*$ for $i$th sample is found by (\ref{eqn:esm_model})
\begin{equation}\label{eqn:esm_model}
v_i^* = \mathbbm{1}_{[\Vec{v}_{esm}^T \cdot \Vec{w}_{esm} >0.5]}
\end{equation}
where $\Vec{v}_{esm} =(v^m)^M_{m=1}$ and  $\Vec{w}_{esm} = (w^m_{esm})^M_{m=1}$.
Therefore, $\mu^m_{\sigma}$ in (\ref{eqn:model_var}) is a sufficient statistic for the ensemble anomaly in (\ref{eqn:esm_model}).
Our proposed estimation algorithm for $\mu^m_{\sigma}$ is inspired by \textit{jackknife+-after-bootstrap (J+aB)} \cite{NEURIPS2020_2b346a0a}.

For bootstrap, our algorithm performs random downsampling of sampling rate $\alpha \in (0.5,1]$, i.e.,  \textit{an ordered sub-sampling with replacement}.
Subsampling randomly selects data index for training  with a fixed sampling ratio $\alpha$ for $B$ bootstrap models. 
The $j$th subsampled data is denoted by $\Tilde{\mathbf{x}}_j$ for $j = 0\dots B$ and $\Tilde{\mathbf{x}}_0 = (t_i, \mathbf{y}_i)_{i = 1\dots N}$ (i.e. the original dataset).
Let $h_{ij}$ denote an indicator such that $h_{ij}=1$ if $i$th sample selected for $j$th bootstrap, and $0$ otherwise for $i=1 \cdots N, j=0\cdots B$.
Assume that $0$th bootstrap uses all training samples (i.e., no subsampling) such that $h_{i0}=1$ for $i = 1\cdots N$.
Then the selected data index is represented as a hot-encoded binary matrix $\mathbf{H}_{N \times B+1}  =[h_{ij}]^{N,B}_{i,j=1,0}$ such that $\sum_i h_{ij}/N \approx \alpha$ for a large $N$ and $B$.
Let define an out-of-sample matrix denoted by $\mathbf{H}^c$ such that $\mathbf{H}^c= [h^c_{ij}]^{N,B}_{i,j=1,0}$ where $h^c_{ij} = |1-h_{ij}|$ indicate an $i$th sample not seen by $j$th bootstrap. We derive a weight matrix  $\mathbf{W}_{N \times B}  =[w_{ij}]^{N,B}_{i,j=1,1}$ where  $w_{ij} = h^c_{ij} / \sum^i h^c_{ij}$ which quantifies the credibility of $j$th bootstrap model for $i$th sample.

Our algorithm consists of two chained functions; embedding function  $f_{emb}$ that learns a regular pattern (i.e., normal data) from unlabelled data to compute dissimilarity of samples from the learnt pattern and probability density function $g_{prob}$ that maps the dissimilarity into probability, that formally described by
$f_{emb} : \mathbf{x}_i \mapsto d_i$ and  $g_{prob} : d_i \mapsto p_i$.
Let $\theta^*_m(j)$ denote the optimal parameter of $m$th embedding model found by solving the optimization $\theta^*_m(j) = \argmin_{\theta \in \Theta_m} \| f^m_{emb}(\mathbf{X}_1^j;\mathbf{\theta}^m)- \mathbf{y}^j \|_2 $
where $\mathbf{X}_1^j = ((t^n, \mathbf{y}_n))^j_{n=1}$.
Let define  $f^m_j\coloneqq f^m_{emb}(\mathbf{x}; \theta^*_m(j) )$ for the learnt embedding function and $\hat{\mathbf{y}}_{ij}^m \coloneqq f^m_j(\mathbf{x}_i)$ for the predicted time series data for $i$th train sample by $j$th bootstrap model. 
Then our algorithm computes a degree of anomaly by
$d^m_{ij} = \| \hat{\mathbf{y}}^m_{ij} - \mathbf{y}_i \|_{2}$, referred to \textit{anomaly distance}, that is the euclidean distance between embedding model prediction from $j$th bootstrap and its residual for $i$th training sample. Our algorithm uses Gaussian Mixture Model (GMM) to learn the probability model of a normal and an abnormal state from the anomaly distance.
Let $g^m_j$ denote GMM model from $j$th bootstrap and $m$th embedding model 
such that $g^m_j: d^m_{ij} \mapsto p^m_{ij}$.
Assuming data contains two classes, the normal and the abnormal state we use GMM with $k=2$ (i.e., two centroids) for the clustering algorithm $g$.
The probability density function of $d^m_{ij}$ for GMM is defined by
$$ g^m_j(x) = \sum_{k=0,1} \pi^m_{kj} N(x | \mu^m_{kj}, \Sigma^m_{kj})$$ 
where $0 \leq \pi_k \leq 1$ is the weight probability  with $~ \sum_k \pi_k=1$ and $N(x|\mu_x, \Sigma_x)$ is Gaussian distribution  of the random variable $x$ with mean $\mu_x$ and covariance matrix $\Sigma_x$. 
The GMM model can be trained by Expectation-Maximization (EM) algorithm with training data $ \{d_{ij} | i=1...n \}$.
Let us assume that $\mu_0 < \mu_1$. Then the anomaly state is voted to $v^m_{ij} =1$ (i.e. abnormal) if $g^m_j(d_{ij}) = 1$, and $0$ otherwise.

Let $r^m_i$ denote the out-of-sample error margin of anomaly probability for $i$th training sample and $m$th embedding model and $\mathbf{r}^m = (r^m_i)^N_{i =1}$.
Let $z^m_i$ denote the weighted average of anomaly probability over all bootstrappings given embedding model $m$. Then it can easily derived that  $z^m_i = <\mathbf{P}^m_{i\cdot} ,\mathbf{W}^m_{i\cdot}>$
where $\mathbf{P}^m_{N\times(B+1)} = [p^m_{ij}]^{N,B}_{i =0, j =0}$, and  $r^m_i = |z^m_i - p^m_{i,0}|$.
Our algorithm is to estimate $\mu^m_{\sigma}$ in (\ref{eqn:model_var}) from anomaly probability samples with out-of-sample residual errors given $\mathbf{X}_{val}$. 
Let us define $q^{m}_{ik} = {z^m_{ki} \pm  r^m_i}$, its mean $\mu^m_k = \mathbbm{E}_i [q^{m}_{ik}]$, and  variance $(\sigma^m_k)^2 = Var_i [q^{m}_{ik}]$ for $k$th validation data.
To estimate $\mu^m_{\sigma}$, our algorithm draws  $L$ samples of  anomaly probability from the distribution (\ref{eqn:proba_sampling}),
\begin{equation}\label{eqn:proba_sampling}
p^m_k \sim \mathcal{N}(\mu^m_k,\sigma^m_k + \epsilon)
\end{equation}
where $\epsilon$ is used to test anomaly model sensitivity on the anomaly threshold $0.5$.
Finally, we can compute the estimated anomaly model variance $\hat{\mu}^m_{\sigma}$ in (\ref{eqn:model_var3})
\begin{equation}\label{eqn:model_var3}
\hat{\mu}^m_{\sigma} = \frac{1}{K} \sum^K_{k=1} \bar{v}^m_k (1-\bar{v}^m_k)
\end{equation}
where $\bar{v}^m_k = \sum^L_l  \mathbbm{1}_{[p >0.5]}(p =p^m_k)$.
                   
The overall algorithm architecture for our anomaly detection algorithm is shown in Fig.\ref{fig:system_a}.
The figure illustrates the individual modules in the bootstrap model pipeline for our proposed method for model 1 to $M$: subsampling, embedding model, Gaussian mixture model, and model evaluation. In the figure, $i$th time series data $\mathbf{y}_i$ comes with its corresponding time stamp $t_i$ for $i = 1\dots N$ where $N$ is the size of unlabelled dataset.

The aforementioned process are done independently by agents in parallel for all combinations of models and  their bootstraps. 
Thus, $M \times B$ agents build a model and cast a vote for an observed sample, then the final voting score is computed their weighted average. 
The final decision on the anomaly is made by aggregating the anomaly probability matrix $\mathbf{P}$ with residual error matrix  $\mathbf{R}$.

\subsection{Boosted Embedding Model}
In order to decrease model variance $\mu^m_{\sigma}$ in (\ref{eqn:model_var}), we leverage embeddings to learn seasonality (e.g., daily, weekly, monthly) or unknown cycles by multiple categorical features. Let's assume $\theta^{T}$ to capture time-categorical features (e.g., months of the year, days of the week, and hours of the day) and $\theta^{I}$ to represent other independent categorical features. Then, the $m-th$ model can be formulated as $f^m := f^m_{emb} (\mathbf{x} ; \theta_m^{T}, \theta_m^{I}) + f^m_{res} (\mathbf{x})$ where $f_{emb}$ is embedding model and $f_{res}$ is a residual model.

We leveraged \textit{DeepGB} algorithm in \cite{karingula2021boosted} to combine gradient boosting with embeddings. In general, gradient boosting trains several simple models sequentially. The key idea of boosting is that each subsequent model trains only on the difference of the output and previous model to leverage each model's strengths and minimize the regression error. Our approach is conducting gradient boosting to fit weak learners on residuals to improve the previous models. We propose a loop wherein, each iteration, we freeze the previous embedding, new embedding added to models, and the network grows. The last residual model can be solved by deep neural networks, SVM, or other approaches. Our approach can be summarized as
$$f^m = [e^m_1, \ldots, e^m_L, r^m].$$
where $e_i, i=1,\cdots, L$ are embedding models to capture categorical data, and $r$ represents the residual model.
As proved in \cite{karingula2021boosted}, the weights of layers can be frozen to simplify the training and skip computing the residuals. The summary of our boosted embedding is presented in \ref{algo:bem}.

\begin{algorithm} [!ht]
	\SetKwInOut{Input}{Input}
	\SetKwInOut{Output}{Output}
	\Input{$\mathbf{X} = ((t_n, \mathbf{y}_n))^N_{n=1}$, $N$: number of samples} 
	\Output{$f^m$}
	
	\emph{$f^m = []$}\\
	\emph{$F_0 := y$}\\
	{For $1\leq l \leq L+1$:} \atcp{iteration over the embedding models} 
	\quad $e_l.fit(t,F_{l-1})$ \atcp{fitting the selected embedding model} 
	\quad $F_l = F_{l-1} - e_l.predict(t)$ \atcp{residual computation}
	\quad \text{if} $|F_l-F_{l-1}|<\epsilon:$ \atcp{termination condition}
	 \qquad \text{break}\\
	\quad $f^m.append(e_l)$\\
	\Return $f^m$
	\caption{Boosted Embeddings Algorithm} \label{algo:bem}
\vspace{-1pt}
\end{algorithm}

\section{Experiment} \label{sec:expr}
We compare our approach, LaF-AD, with state-of-the-art anomaly detection techniques in the literature: 1) KNN~\cite{Fabrizio2002}, 2) Isolation Forest (I.F)~\cite{liu2008isolation}, and 3) an LSTM based Autoencoder (AeLSTM)~\cite{malhotra2016lstm}. First, we evaluate this comparison on a synthetic dataset with injected anomalies (Section~\ref{sec:case_panw_syn_data}). Second, we conduct experiment on a public dataset with a benchmark anomaly detection data set (Section~\ref{sec:case_public}). 

\noindent\textbf{Implementation Detail:}
All algorithms are implemented in \texttt{Python 3.7}. Methods DeepGB, AeLSTM and LaF-AD are implemented using \texttt{TensorFlow 2.4.0}, while the remaining approaches, i.e., IF, and KNN are implemented using \texttt{scikit-learn 0.24.1}. Experiments are performed on MacBook Pro with 12-core CPUs, 16 GB RAM.

\noindent\textbf{Experimental Protocol:}
We apply the Algorithm.\ref{fig:system_a} as is, since the selected public datasets are provided as preprocessed. Further, we enlarge the feature space by creating a rolling window of size $W$ for each timestamp. For example, if $W=24$, then each timestamp will have the previous $24$ values as features.
In our setting we employ  $80\% $ data for training and the remaining for validation $5$ times. 

\begin{table}
	\centering
	\caption{Default parameter setting} \label{tbl:param_setting}
\begin{center}
\begin{tabular}{cc}
\hline
\multicolumn{2}{c}{Algorithm} \\
I.F & \texttt{KNN} \\
\cmidrule(lr) {1-2} $t: 100$ & $n: 5$\\
- & $\psi :256$ $c: 0.1$ \\\hline
\end{tabular}
\end{center}
\begin{tablenotes}
		\small
		\centering
		\item  $t$:the number of trees,  $\psi$: sub-sampling size,  $k$:the number of neighbors,
		$\kappa $: kernel function,  $f_{rbf} (x,y)= e^{-\|x-y\|^2/c}$, $\nu$:outliers fraction,
		c:contamination, n:\#neighbors
	\end{tablenotes}
\end{table}


Table $\ref{tbl:param_setting}$ summarizes the default parameter settings for our experiment in this case study. 
We use the default optimal settings for I.F recommended in their original papers \cite{Liu12,Breunig00}.

For KNN, there is no specific recommended default parameter setting. Instead, their optimal parameters need to be tuned by evaluating out-of-sample errors via cross-validation during the training.
It, however, is not feasible in our problem setting and many industrial applications as no labels are available during at the time of training. A number of experiments has been conducted with randomly selected parameters for KNN, but we was not able to find any meaningful changes in performance.  Hence, we use a default parameter setting for KNN given by  scikit-learn package \cite{scikit-learn}.  For AE-LSTM, numerous possible structures with different performances are possibles. We started with the architecture in Malhotra et al.~\cite{malhotra2016lstm} and fine tuned to the most exemplary structure that gives one of the best AUC performances.  

\subsection{Simulating synthetic anomaly data}\label{sec:case_panw_syn_data}
We now analyze the performance of the algorithm on a synthetic dataset with simulated noise and injected anomalies. The goal of this experiment is to build a relationship between the signal-to-noise ratio (SNR) of the data and our method's performance in a controlled environment. The synthetic data is generated from the following structural time series model with additive components:
\begin{gather*}
    y_t = x_t + z_t + \epsilon_t
\end{gather*}
Where $x_t$ is a periodic signal, $z_t$ is a noise component, and $\epsilon_t$ are injected anomalies. More specifically:
\begin{gather*}
    x_t = a\sin(2\pi ft + \phi) + b \\
    z_t = \phi_1 y_{t-1} + \phi_2y_{t-2} + \theta_1w_t + \theta_2w_{t-2}\\
    = \mathrm{ARMA}(2,2), w_t \sim \mathrm{WhiteNoise(0, \sigma_w^2)}
\end{gather*}
Where the parameters for $x_t$ are:
\begin{gather*}
   a=10, b=20, \phi=0, f = 5/T, T=240
\end{gather*}
where $T$ is the total number of time steps in a 5-day period at 30-minute increments. Similarly, $z_t$ is determined by:
\begin{gather}
    \phi_1 = \frac{1}{2}, \phi_2 = -\frac{1}{2}, \theta_1 = \theta_2 = 2, \sigma_w^2=1
\end{gather}
And finally, anomalies are injected through a mixture distribution $\epsilon_t$ which follows:
\begin{equation}
  \epsilon_t \stackrel{iid}{\sim}
    \begin{cases}
      \mathrm{Poisson}(\lambda_1) & \text{with probability $\pi$}\\
       c_{min} + \mathrm{Poisson}(\lambda_2) & \text{with probability $1 - \pi$}
    \end{cases}       
\end{equation}
With the $\epsilon_t$'s parameters set to the values:
\begin{gather*}
    \pi = 0.999, \lambda_1 < \lambda_2, \lambda_1 = c_{min} = 10
\end{gather*}
which are chosen to ensure a sparse ($\mathbb E[\text{\# of Anomalies}]=n/1000$) yet realistic amount of injected anomalies. To better simulate realistic patterns of life, we alternate between generating enough observations from $y_t$ for a 5-day work week followed by two days of weekend behavior by setting $x_t = z_t = c_{weekend} =0$.
We proceed by setting $n=13,497$ and varying the window size and $\lambda_2/\lambda_1$ to analyze the performance of our algorithm. The results of this simulation experiment are summarized in Table \ref{tab:simulation}.\\


\begin{table}[!ht]
	\centering
	\caption{AUC  Performance summary Simulation Experiment for $\lambda_2/\lambda_1=1$} \label{tab:simulation}
\begin{tabular}{cccccc}
\hline
Dataset & $W$ & \texttt{LaF-AD} & \texttt{I.F} & \texttt{KNN} & \texttt{AeLSTM}\\
\cmidrule(lr) {1-2} \cmidrule(lr) {3-6}  
\texttt{Simluation 1} & 5points & 0.987 & 0.712 & 0.821 & 0.987\\
\cmidrule(lr) {1-2} \cmidrule(lr) {3-6}      
Variance & - & 0.0012 & 0.0032 & 0.0021 & 0.0012\\ \hline
\end{tabular}
\end{table}

\subsection{ Case Study 2 : Validation with Public Datasets}\label{sec:case_public}
We now evaluate our methodology on a public data set taken from the Numenta Anomaly Benchmark \footnote{NAB data can be retrieved here \url{https://github.com/numenta/NAB/tree/master/data}} (NAB) repository \cite{AHMAD2017134}. Specifically, we investigate the dataset 
\texttt{nyc taxi} (\texttt{taxi}) which describes the total number of taxi passengers in New York City taken from the NYC Taxi and Limousine Commission. The data is count-type data with non-negative integer support. The dataset consists of 5 anomalies occuring at various holidays and severe weather events.

\texttt{taxi} contains timestamps measured at regular intervals and provides a univariate time series for experimentation. Since the data set has labels, this enables us to compute various accuracy metrics on the known anomaly timestamps. \texttt{taxi} is typical of most telemetry datasets seen in practice; where sample size is large relative to a sparse set of anomalies.  A numerical summary of the data set is presented in Table \ref{tbl:public} along with results in Table \ref{tbl:auc_final}:
\begin{table}[ht]
	\caption{Public Datasets}
	\label{tbl:public}
	\begin{subfigure}[b]{0.8\textwidth}
	\begin{center}
	\begin{tabular}{cccccc}
		\toprule
		Dataset    & Count  &Start Date  & End Date  & Frequency & Number of Anomalies \\
		\cmidrule(lr) {1-6} 
		\texttt{nyc\_taxi} & 10320  & 2014-07-01 & 2015-01-31 & 30 minutes & 5 \\
		\bottomrule
	\end{tabular}
	\end{center}
	\end{subfigure}
	
\end{table}

\begin{table}[!ht]
	\centering
	\caption{AUC  Performance summary for \texttt{taxi} dataset} \label{tbl:auc_final}
\begin{tabular}{cccccc}
\hline
Dataset & $W$ & \texttt{LaF-AD} & \texttt{I.F} & \texttt{KNN} & \texttt{AeLSTM}\\
\cmidrule(lr) {1-2} \cmidrule(lr) {3-6}  
\texttt{taxi} & 5points & 0.82 & 0.44 & 0.82 & 0.53\\
\cmidrule(lr) {1-2} \cmidrule(lr) {3-6}      
Variance & - & 0.0023 & 0.0041 & 0.0023 & 0.0042\\ \hline
\end{tabular}
\end{table}

In Table \ref{tbl:auc_final}, the AUC results are summarized for other comparable baseline algorithms.  The best performance is marked with bold font. As we observe, \texttt{LaF-AD} selects the best model compared to other baselines. We can conclude that lower model variance is the indicator of high AUC without knowing the labels.  

The table shows the experiment result of \texttt{LaF-AD} with other baseline anomaly detection algorithms for different window sizes. For the experiment, we assume the label is not available before the deployment, hence it is unknown to us which baseline algorithm and window size offers the best or the worst performance.  It can easily seen that our algorithm consistently can outperforms  the best baseline algorithm for all window sizes. 
The average and standard deviation of AUC performance over window sizes for each algorithm are shown in the last two rows. 
It shows that our algorithm not only outperforms in the average AUC  but also has the most stable performance (i.e., the smallest model variance learned by our algorithm).

\section{Conclusion} \label{sec:conclude}
Unsupervised anomaly detection methods are widely used in a variety of research areas. In this paper, we propose a novel label-free anomaly detection algorithm, LaF-AD, for time series data. Furthermore, we develop a model evaluation metric based on the variance that quantifies the sensitivity of anomaly probability by learning bootstrapped models. We derive a new performance bound for bootstrap prediction. Empirical evaluations on both synthetic and public benchmark datasets demonstrate that the proposed method outperforms state-of-the-art unsupervised anomaly detection models for univariate time series. 
Finally, this paper opens up several new directions for further research. Extensive evaluation of our method on other complex domains is an immediate direction. 
The current approach is designed and model with univariate time-series analysis, However extending it to multivariate time series is an open problem and interesting to us from a practical application standpoint.
\vspace{12pt}
\bibliographystyle{authordate1}
\bibliography{panw_anomaly}

\begin{thebibliography}{}

\bibitem[\protect\citename{Aggarwal, }2016]{Aggarwal16}
Aggarwal, Charu~C. 2016.
\newblock Outlier Analysis.
\newblock 2nd edn.
\newblock Springer Publishing Company.

\bibitem[\protect\citename{Ahmad {\em et~al.}, }2017]{AHMAD2017134}
Ahmad, Subutai, Lavin, Alexander, Purdy, Scott, \& Agha, Zuha. 2017.
\newblock Unsupervised real-time anomaly detection for streaming data.
\newblock {\em Neurocomputing}, {\bf 262}, 134--147.
\newblock Online Real-Time Learning Strategies for Data Streams.

\bibitem[\protect\citename{Angiulli \& Pizzuti, }2002]{Fabrizio2002}
Angiulli, Fabrizio, \& Pizzuti, Clara. 2002.
\newblock Fast Outlier Detection in High Dimensional Spaces.
\newblock {\em Pages  15--26 of:} Elomaa, Tapio, Mannila, Heikki, \& Toivonen,
  Hannu (eds), {\em Principles of Data Mining and Knowledge Discovery, 6th
  European Conference, {PKDD} 2002, Helsinki, Finland, August 19-23, 2002,
  Proceedings},  vol. 2431.
\newblock Springer.

\bibitem[\protect\citename{Braei \& Wagner, }2020]{braei2020anomaly}
Braei, Mohammad, \& Wagner, Sebastian. 2020.
\newblock Anomaly detection in univariate time-series: A survey on the
  state-of-the-art.
\newblock {\em arXiv preprint arXiv:2004.00433}.

\bibitem[\protect\citename{Breunig {\em et~al.}, }2000a]{breunig2000lof}
Breunig, Markus~M, Kriegel, Hans-Peter, Ng, Raymond~T, \& Sander, J{\"o}rg.
  2000a.
\newblock LOF: identifying density-based local outliers.
\newblock {\em Pages  93--104 of:} {\em Proceedings of the 2000 ACM SIGMOD
  international conference on Management of data}.

\bibitem[\protect\citename{Breunig {\em et~al.}, }2000b]{Breunig00}
Breunig, Markus~M., Kriegel, Hans-Peter, Ng, Raymond~T., \& Sander, J\"{o}rg.
  2000b.
\newblock LOF: Identifying Density-based Local Outliers.
\newblock {\em In:} {\em Proceedings of the 2000 ACM SIGMOD International
  Conference on Management of Data}.
\newblock SIGMOD '00.

\bibitem[\protect\citename{Cai {\em et~al.}, }2015]{Yongjie2015}
Cai, Yongjie, Tong, Hanghang, Fan, Wei, Ji, Ping, \& He, Qing. 2015.
\newblock Facets: Fast Comprehensive Mining of Coevolving High-Order Time
  Series.
\newblock {\em In:} {\em Proceedings of the 21th ACM SIGKDD International
  Conference on Knowledge Discovery and Data Mining}.
\newblock New York, NY, USA: Association for Computing Machinery.

\bibitem[\protect\citename{Campos {\em et~al.}, }2016]{campos2016evaluation}
Campos, Guilherme~O, Zimek, Arthur, Sander, J{\"o}rg, Campello, Ricardo~JGB,
  Micenkov{\'a}, Barbora, Schubert, Erich, Assent, Ira, \& Houle, Michael~E.
  2016.
\newblock On the evaluation of unsupervised outlier detection: measures,
  datasets, and an empirical study.
\newblock {\em Data mining and knowledge discovery}, {\bf 30}(4), 891--927.

\bibitem[\protect\citename{Chen {\em et~al.}, }2017]{chen2017outlier}
Chen, Jinghui, Sathe, Saket, Aggarwal, Charu, \& Turaga, Deepak. 2017.
\newblock Outlier detection with autoencoder ensembles.
\newblock {\em Pages  90--98 of:} {\em Proceedings of the 2017 SIAM
  international conference on data mining}.
\newblock SIAM.

\bibitem[\protect\citename{Cho {\em et~al.}, }2014]{cho2014learning}
Cho, Kyunghyun, Van~Merri{\"e}nboer, Bart, Gulcehre, Caglar, Bahdanau, Dzmitry,
  Bougares, Fethi, Schwenk, Holger, \& Bengio, Yoshua. 2014.
\newblock Learning phrase representations using RNN encoder-decoder for
  statistical machine translation.
\newblock {\em arXiv preprint arXiv:1406.1078}.

\bibitem[\protect\citename{Dempster {\em et~al.}, }1977]{Dempster1977}
Dempster, A.~P., Laird, N.~M., \& Rubin, D.~B. 1977.
\newblock Maximum Likelihood from Incomplete Data via the EM Algorithm.
\newblock {\em Journal of the Royal Statistical Society. Series B
  (Methodological)}.

\bibitem[\protect\citename{Goernitz {\em et~al.}, }2015]{Goernitz2015}
Goernitz, Nico, Braun, Mikio, \& Kloft, Marius. 2015.
\newblock Hidden Markov Anomaly Detection.
\newblock {\em Pages  1833--1842 of:} Bach, Francis, \& Blei, David (eds), {\em
  Proceedings of the 32nd International Conference on Machine Learning}.
\newblock Proceedings of Machine Learning Research, vol. 37.
\newblock Lille, France: PMLR.

\bibitem[\protect\citename{G{\"{o}}rnitz {\em et~al.}, }2014]{GornitzKRB14}
G{\"{o}}rnitz, Nico, Kloft, Marius, Rieck, Konrad, \& Brefeld, Ulf. 2014.
\newblock Toward Supervised Anomaly Detection.
\newblock {\em CoRR}.

\bibitem[\protect\citename{Hariri {\em et~al.}, }2021]{Hariri_2021}
Hariri, Sahand, Kind, Matias~Carrasco, \& Brunner, Robert~J. 2021.
\newblock Extended Isolation Forest.
\newblock {\em IEEE Transactions on Knowledge and Data Engineering}, {\bf
  33}(Apr), 1479–1489.

\bibitem[\protect\citename{Hoffmann, }2007]{hoffmann2007kernel}
Hoffmann, Heiko. 2007.
\newblock Kernel PCA for novelty detection.
\newblock {\em Pattern recognition}, {\bf 40}(3), 863--874.

\bibitem[\protect\citename{Karingula {\em et~al.}, }2021]{karingula2021boosted}
Karingula, Sankeerth~Rao, Ramanan, Nandini, Tahsambi, Rasool, Amjadi, Mehrnaz,
  Jung, Deokwoo, Si, Ricky, Thimmisetty, Charanraj, \& Coelho~Jr,
  Claudionor~Nunes. 2021.
\newblock Boosted Embeddings for Time Series Forecasting.
\newblock {\em arXiv preprint arXiv:2104.04781}.

\bibitem[\protect\citename{Kim {\em et~al.}, }2020]{NEURIPS2020_2b346a0a}
Kim, Byol, Xu, Chen, \& Barber, Rina. 2020.
\newblock Predictive inference is free with the jackknife+-after-bootstrap.
\newblock {\em Pages  4138--4149 of:} Larochelle, H., Ranzato, M., Hadsell, R.,
  Balcan, M.~F., \& Lin, H. (eds), {\em Advances in Neural Information
  Processing Systems},  vol. 33.
\newblock Curran Associates, Inc.

\bibitem[\protect\citename{Kirschen {\em et~al.}, }2003]{kirschen2003computing}
Kirschen, DS, Bell, KRW, Nedic, DP, Jayaweera, D, \& Allan, RN. 2003.
\newblock Computing the value of security.
\newblock {\em IEE Proceedings-Generation, Transmission and Distribution}, {\bf
  150}(6), 673--678.

\bibitem[\protect\citename{Knorr {\em et~al.}, }2000]{knorr2000}
Knorr, Edwin~M, Ng, Raymond~T, \& Tucakov, Vladimir. 2000.
\newblock Distance-based outliers: algorithms and applications.
\newblock {\em The VLDB Journal}, {\bf 8}(3), 237--253.

\bibitem[\protect\citename{Li {\em et~al.}, }2009]{Junlei2009}
Li, Junlei, McCann, James, Pollard, Nancy, \& Faloutsos, Christos. 2009.
\newblock DynaMMo: Mining and Summarization of Coevolving Sequences with
  Missing Values.
\newblock {\em In:} {\em Proceedings of 15th ACM SIGKDD International
  Conference on Knowledge Discovery and Data Mining (KDD '09)}.

\bibitem[\protect\citename{Liu {\em et~al.}, }2008]{liu2008isolation}
Liu, Fei~Tony, Ting, Kai~Ming, \& Zhou, Zhi-Hua. 2008.
\newblock Isolation forest.
\newblock {\em Pages  413--422 of:} {\em 2008 eighth ieee international
  conference on data mining}.
\newblock IEEE.

\bibitem[\protect\citename{Liu {\em et~al.}, }2012a]{liu2012isolation}
Liu, Fei~Tony, Ting, Kai~Ming, \& Zhou, Zhi-Hua. 2012a.
\newblock Isolation-based anomaly detection.
\newblock {\em ACM Transactions on Knowledge Discovery from Data (TKDD)}, {\bf
  6}(1), 3.

\bibitem[\protect\citename{Liu {\em et~al.}, }2012b]{Liu12}
Liu, Fei~Tony, Ting, Kai~Ming, \& Zhou, Zhi-Hua. 2012b.
\newblock Isolation-Based Anomaly Detection.
\newblock {\em ACM Trans. Knowl. Discov. Data}, {\bf 6}(1).

\bibitem[\protect\citename{Malhotra {\em et~al.}, }2016]{malhotra2016lstm}
Malhotra, Pankaj, Ramakrishnan, Anusha, Anand, Gaurangi, Vig, Lovekesh,
  Agarwal, Puneet, \& Shroff, Gautam. 2016.
\newblock LSTM-based encoder-decoder for multi-sensor anomaly detection.
\newblock {\em arXiv preprint arXiv:1607.00148}.

\bibitem[\protect\citename{Pedregosa {\em et~al.}, }2011]{scikit-learn}
Pedregosa, F., Varoquaux, G., Gramfort, A., Michel, V., Thirion, B., Grisel,
  O., Blondel, M., Prettenhofer, P., Weiss, R., Dubourg, V., Vanderplas, J.,
  Passos, A., Cournapeau, D., Brucher, M., Perrot, M., \& Duchesnay, E. 2011.
\newblock Scikit-learn: Machine Learning in {P}ython.
\newblock {\em Journal of Machine Learning Research}, {\bf 12}, 2825--2830.

\bibitem[\protect\citename{Principi {\em et~al.}, }2017]{principi2017acoustic}
Principi, Emanuele, Vesperini, Fabio, Squartini, Stefano, \& Piazza, Francesco.
  2017.
\newblock Acoustic novelty detection with adversarial autoencoders.
\newblock {\em Pages  3324--3330 of:} {\em 2017 International Joint Conference
  on Neural Networks (IJCNN)}.
\newblock IEEE.

\bibitem[\protect\citename{Ramaswamy {\em et~al.},
  }2000]{ramaswamy2000efficient}
Ramaswamy, Sridhar, Rastogi, Rajeev, \& Shim, Kyuseok. 2000.
\newblock Efficient algorithms for mining outliers from large data sets.
\newblock {\em Pages  427--438 of:} {\em Proceedings of the 2000 ACM SIGMOD
  international conference on Management of data}.

\bibitem[\protect\citename{Rousseeuw \& Hubert, }2018]{rousseeuw2018anomaly}
Rousseeuw, Peter~J, \& Hubert, Mia. 2018.
\newblock Anomaly detection by robust statistics.
\newblock {\em Wiley Interdisciplinary Reviews: Data Mining and Knowledge
  Discovery}, {\bf 8}(2), e1236.

\bibitem[\protect\citename{Sch{\"o}lkopf {\em et~al.}, }2001]{scholkopf2001}
Sch{\"o}lkopf, Bernhard, Platt, John~C, Shawe-Taylor, John, Smola, Alex~J, \&
  Williamson, Robert~C. 2001.
\newblock Estimating the support of a high-dimensional distribution.
\newblock {\em Neural computation}, {\bf 13}(7), 1443--1471.

\bibitem[\protect\citename{Sindhwani {\em et~al.}, }2005]{sindhwani2005beyond}
Sindhwani, Vikas, Niyogi, Partha, \& Belkin, Mikhail. 2005.
\newblock Beyond the point cloud: from transductive to semi-supervised
  learning.
\newblock {\em Pages  824--831 of:} {\em Proceedings of the 22nd international
  conference on Machine learning}.

\bibitem[\protect\citename{Singh \& Chatterjee, }2017]{singh2017cloud}
Singh, Ashish, \& Chatterjee, Kakali. 2017.
\newblock Cloud security issues and challenges: A survey.
\newblock {\em Journal of Network and Computer Applications}, {\bf 79},
  88--115.

\bibitem[\protect\citename{Song {\em et~al.}, }2007]{song2007}
Song, Xiuyao, Wu, Mingxi, Jermaine, Christopher, \& Ranka, Sanjay. 2007.
\newblock Conditional Anomaly Detection.
\newblock {\em IEEE Transactions on Knowledge and Data Engineering}, {\bf
  19}(5), 631--645.

\bibitem[\protect\citename{Vapnik, }1999]{vapnik1999overview}
Vapnik, Vladimir~N. 1999.
\newblock An overview of statistical learning theory.
\newblock {\em IEEE transactions on neural networks}, {\bf 10}(5), 988--999.

\bibitem[\protect\citename{Wen {\em et~al.}, }2017]{wen2017new}
Wen, Long, Gao, Liang, \& Li, Xinyu. 2017.
\newblock A new deep transfer learning based on sparse auto-encoder for fault
  diagnosis.
\newblock {\em IEEE Transactions on Systems, Man, and Cybernetics: Systems},
  {\bf 49}(1), 136--144.

\bibitem[\protect\citename{Xiong {\em et~al.}, }2011]{Liang2011}
Xiong, Liang, Chen, Xi, \& Schneider, Jeff. 2011.
\newblock Direct Robust Matrix Factorizatoin for Anomaly Detection.
\newblock {\em Pages  844--853 of:} {\em 2011 IEEE 11th International
  Conference on Data Mining}.

\bibitem[\protect\citename{Xu {\em et~al.}, }2019]{Xu_2019}
Xu, Zekun, Kakde, Deovrat, \& Chaudhuri, Arin. 2019.
\newblock Automatic Hyperparameter Tuning Method for Local Outlier Factor, with
  Applications to Anomaly Detection.
\newblock {\em 2019 IEEE International Conference on Big Data (Big Data)}, Dec.

\end{thebibliography}

\end{document}